\documentclass[letterpaper, 10pt,  journal, final, twocolumn]{IEEEtran}
\usepackage{graphicx}
\usepackage{amsmath,amssymb}
\usepackage{color}
\usepackage{balance}
\usepackage{mathptmx}
\usepackage[mathscr]{euscript}
\usepackage{amsthm}
\usepackage{cases}
\usepackage{multirow}
\usepackage{setspace}
\usepackage{empheq}
\usepackage{caption}
\usepackage{notoccite}

\usepackage{booktabs}

 \newcommand{\Real}{\mathbb{R}}

 \newcommand{\blue}{\color{blue}}

\begin{document}

\title{{\small\textbf{\blue IEEE/MTS Global Oceans 2020, Singapore-US Gulf Coast}}\\
\Huge {T$^\star$-Lite: A Fast Time-Risk Optimal Motion Planning Algorithm for Multi-Speed Autonomous Vehicles}
\thanks{$^\dag$Dept. of Electrical and Computer Engineering, University of Connecticut, Storrs, CT 06269, USA.}
\thanks{$^\ddag$Naval Undersea Warfare Center, Newport, RI 02841, USA.}
\thanks{$^{\star}$ Corresponding Author (email id: james.wilson@uconn.edu)}
}
\author{ \begin{tabular}{cccc}
{James P. Wilson$^\dag$$^\star$} & {Zongyuan Shen$^\dag$} & {Shalabh Gupta$^\dag$} & {Thomas A. Wettergren$^\ddag$}\\
\end{tabular}
\thanks{This work was supported by US Office of Naval Research under Award Number N000141613032. Any opinions or findings herein are those of the authors and do not necessarily reflect the views of the sponsoring agencies.}
\\ \vspace{-6pt}
}

\maketitle

\begin{abstract}
    In this paper, we develop a new algorithm, called T$^\star$-Lite, that enables fast time-risk optimal motion planning for variable-speed autonomous vehicles. The T$^\star$-Lite algorithm is a significantly faster version of the previously developed T$^\star$ algorithm. T$^\star$-Lite uses the novel time-risk cost function of T$^\star$; however, instead of a grid-based approach, it uses an asymptotically optimal sampling-based motion planner. Furthermore, it utilizes the recently developed Generalized Multi-speed Dubins Motion-model (GMDM) for sample-to-sample kinodynamic motion planning. The sample-based approach  and GMDM significantly reduce the computational burden of T$^\star$ while providing reasonable solution quality. The sample points are drawn from a four-dimensional configuration space consisting of two position coordinates plus vehicle heading and speed. Specifically, T$^\star$-Lite enables the motion planner to select the vehicle speed and direction based on its proximity to the obstacle to generate faster and safer paths. In this paper, T$^\star$-Lite is developed using the RRT$^*$ motion planner, but adaptation to other motion planners is straightforward and depends on the needs of the planner.
\end{abstract}


\thispagestyle{empty}

\section{Introduction}
\label{sec:intro}
Autonomous vehicles are becoming increasingly useful and cost-effective for a variety of tasks in many scientific expeditions. For example, unmanned air vehicles (UAVs) are used for survey, monitoring and mapping~\cite{dunbabin2012robots}. On the other hand, unmanned underwater vehicles (UUVs) are used for exploration~\cite{song2018varepsilon}, oceanic data collection (e.g., salinity and temperature), seabed mapping~\cite{palomeras2018, shen2016,Shen2017,shen2019}, oil spill cleaning~\cite{SGH13}, and mine hunting~\cite{MGR11}. Despite recent advances, the autonomy of these vehicles is limited, especially for curvature-constrained vehicles that operate in environments with many obstacles where finding the time-optimal path is difficult, and is in fact NP-Hard \cite{NPHard1}. At the same time, it is also important to consider the vehicle safety by generating robust collision-free paths. Finally, the missions of autonomous vehicles might require on-demand path synthesis in dynamic environments with currents~\cite{bakolas2013,song2019rapid,MG19} or moving targets. It is thus of practical importance for motion planners to construct \textit{time-risk optimal} paths for autonomous vehicles rapidly. 

Recent research has focused on finding approximate shortest paths in obstacle-rich environments for single speed non-holonomic vehicles; a review is presented in \cite{ZENG2015_AUVPathPlanningSurvey}. 
Sample-based methods based on probabilistic road maps (PRM) \cite{Kavraki1996_PRM} and rapidly-exploring random trees (RRT) \cite{LaValle1999_RRT} are becoming increasingly popular for on-demand motion planning because they can find feasible solutions quickly in high-dimensional spaces. Of particular note are PRM$^*$ and RRT$^*$ \cite{Karaman2011_PRMRRTStar} since they provide asymptotically optimal solutions as the number of sampled way points increases. 
However, the above methods are restricted to single speed vehicles, thus they can generate only shortest paths not time-optimal paths.

Recent research has also developed motion planners that focus on vehicle safety when traversing in obstacle-rich environments~\cite{Hernandez2016_Saftey2,Liu2017_Saftey3}. 
In general, robust collision-free paths are achieved by creating a buffer around the autonomous vehicle and obstacles. However, these methods do not consider the full state of the vehicle (i.e., its position, heading, and speed) in relation to the obstacles for computing the collision risk. Furthermore, some of these methods do not consider travel time costs, thus generating longer paths. 

To the best of our knowledge, our recently developed T$^\star$ algorithm \cite{Song2019_TStar} is the only motion planner for multi-speed vehicles that gives time-optimal risk-aware paths by considering both time and risk costs together. In T$^\star$, the risk cost of a set of candidate paths is determined by estimating the collision time of the vehicle with an obstacle tangent to its current trajectory direction and using its current speed. The time cost is computed from the path segment lengths and velocities. Then, a joint cost function is formulated considering both risk and time. Finally, the time-risk optimal path is obtained by performing A$^*$-like search in a high-dimensional discrete configuration space considering vehicle poses. While T$^\star$ indeed provides the near optimal time-risk solution, its computation cost limits its feasibility in real-time applications.

In all these motion planners, the solution quality depends on the underlying kinodynamic \textit{motion model} used to connect way points. In particular, the Dubins motion model is used for single forward velocity vehicles with a minimum turning radius. Dubins \cite{D57} provides the minimum-time paths in open areas for such vehicles. The minimum-time path is one from six path types composed of $L\equiv$left turn, $S\equiv$straight line, and $R\equiv$right turn segments. Reeds-Shepp curves \cite{reedsshepp} extend Dubins curves by also considering the backward velocity. While these paths are fast to compute, they provide the sub-optimal travel time and risk costs for variable-speed vehicles due to the single-speed limitation. Recently, Wolek, et al. \cite{WCW16} proposed the time-optimal solution for variable-speed vehicles using the minimum and maximum speeds, but the high computational cost of their model makes it infeasible for real-time applications. Even when used offline, Wolek's motion model yields sub-optimal results for joint time-risk planning since the model does not considers the risk. Furthermore, the straight segments of paths always operate at max speed, which greatly increases the collision risk. 

As such, the standard motion models for variable-speed vehicles in literature are insufficient for online time-risk motion planning. An ideal desired motion model would enable: i) better maneuverability by controlling the vehicle speed and thus turning radius, ii) risk mitigation by selecting speeds and headings based on obstacle proximity, and iii) fast computation for real-time application. To the best of our knowledge, the only motion model with these features is our recently developed Generalized Multi-speed Dubins Motion-model (GMDM) for variable-speed vehicles \cite{Wilson2019_GDubins}. Unlike other models, GMDM achieves superior time-risk costs in T$^\star$ while having a low computational complexity. This is achieved by extending the Dubins model to allow each path segment to have an optimized speed for time-risk optimal motion planning.

In lieu of the above discussion and limitations of the existing literature, we develop a new algorithm in this paper, called T$^\star$-Lite, which enables fast time-risk optimal motion planning for variable-speed vehicles. This is achieved by: i) porting the time-risk cost function from T$^\star$ into an asymptotically optimal sampling-based motion planner and ii) utilizing the Generalized Multi-speed Dubins Motion-model for point to point motion planning. The sample-based approach significantly reduces the computational overhead of T$^\star$ while providing reasonable solution quality. In particular, the sampled points are drawn from a four-dimensional configuration space consisting of two-dimensional position coordinates, and vehicle heading and speed. At the same time, GMDM enables exploitation of the time-risk cost function to yield near-optimal multi-speed paths connecting the sampled points. Specifically, T$^\star$-Lite enables the motion planner to select the vehicle speed and direction based on the proximity to the obstacle to allow for faster and safer paths. In order to improve the convergence rate and provide higher quality solutions under computation time budgets, smart pruning techniques will be developed in future work to better select heading and speed selection of generated way points based on the collision time with nearby obstacles. Without loss of generality, T$^\star$-Lite is initially developed in this paper using the RRT$^*$ motion planner. However, adaptation to other motion planners is straightforward and depends on the user's needs and mission specifications.

The rest of this paper is organized as follows. Section~\ref{sec:probform} formulates the time-risk optimal motion planning problem for variable-speed autonomous vehicles. Section~\ref{sec:algorithm} presents the details of the T$^\star$-Lite algorithm. Section~\ref{sec:results} presents the results on a simulated scenario, and Section~\ref{sec:conclusions} concludes this paper with recommendations for future work.

\section{Problem Formulation}
\label{sec:probform}

Let $A\subset\Real^2$ be a 2D search area with obstacles. Consider an inertial vehicle in this plane. The vehicle motion is:
\begin{subequations}
\label{eq:vehiclemotion}
\begin{align}
    \dot{x}(t) &= v(t)\cos{\theta(t)}\\
    \dot{y}(t) &= v(t)\sin{\theta(t)}\\
    \dot{\theta}(t) &= u(t)
\end{align}
\end{subequations}
where the $(x,y,\theta)\in SE(2)$, $u(t)$ is the turning rate, and $v(t)$ is the vehicle speed. Specifically, the turning rate $u(t)\in [-u_{max},u_{max}]$ is symmetric and bounded, where $u_{max}\in\Real^+$ and $+/-$ refers to a left/right turn. Without loss of generality, $v(t)\in [v_{min},v_{max}]$ where $v_{min},v_{max}\in\Real^+$ are the minimum and maximum speeds of the vehicle, respectively. 

The speed $v(t)$ and turning rate $u(t)$ are connected by the vehicle curvature $\kappa(t)=u(t)/v(t)$. As such, the curvature is bounded by $0\leq |\kappa(t)|\leq u_{max}/v_{min}$. When $\kappa(t)=0$, the vehicle is moving forward on a straight line. When $\kappa(t)=|u_{max}|/v_{min}$ or $\kappa(t)=|u_{max}|/v_{max}$, the vehicle is moving forward along a curve at either the minimum or maximum turning radius, respectively, where the vehicle turning radius $r(t)=1/\kappa(t)$.

A vehicle state is a four-dimensional vector defined as $\mathbf{p}=(x,y,\theta,v)$. Let $\Gamma$ be the set of collision-free paths between the start state $\mathbf{p}_{start}$ and goal state $\mathbf{p}_{goal}$. For each path $\gamma\in\Gamma$, the control $\mathbf{c}(s)=(\kappa,v)$ at any point $s$ on $\gamma$ belongs to the following constraint set \cite{WCW16}:

\begin{equation}
    \label{eq:pathconstraint}
    \Omega = \Big\{(\kappa,v)\mid v_{min}\leq v\leq v_{max} \ \textrm{and} \ |\kappa|\leq \frac{u_{max}}{v}\Big\}
\end{equation}

The admissible control must satisfy the boundary conditions of the search area and be piece-wise continuous, i.e., $\mathbf{c}$ must drive the vehicle from $\mathbf{p}_{start}$ to $\mathbf{p}_{goal}$ along some feasible path $\gamma$. Let $\mathscr{R}(s)$ denote the risk at point $s$ along path $\gamma$. Then, the time-risk cost of a feasible path $\gamma$ is defined as:
\begin{equation}
\label{eq:totalcost}
    J(\gamma) = \int_\gamma \mathscr{R}(s)\cdot \frac{1}{v(s)} ds
\end{equation}
where $1/v(s)$ evaluates the time cost along a segment $ds$.

Therefore, the objective is to find the control $\mathbf{c}^* \in \Omega$ which generates the best feasible path $\gamma^*$ that yields the minimal cost $J(\gamma^*)$ such that $J(\gamma^*)\leq J(\gamma)\;\forall\gamma\in\Gamma$.

\begin{figure*}[t!]
    \centering
        \centering
        \includegraphics[width=0.8\textwidth]{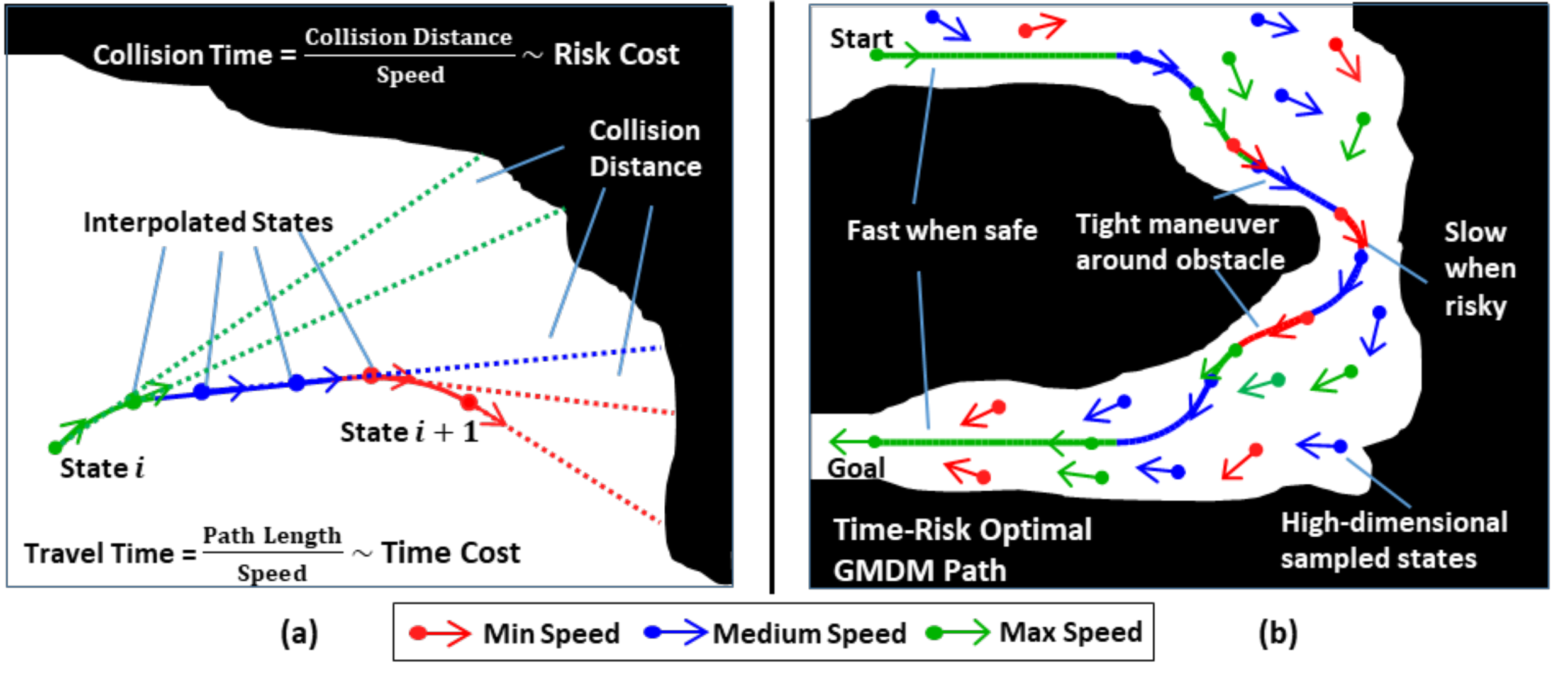}
         \caption{Visualization of the main features of T$^\star$-Lite. Fig. 1 (a) provides an overview of the computation of time and risk costs in the joint optimization problem. Fig. 1 (b) shows an example of the high-dimensional sampled way points of the vehicle states and the time-risk optimal solution generated from RRT$^*$ using the Generalized Multi-speed Dubins Motion model.}
         \label{fig:Figure1}
\end{figure*}

\section{T$^\star$-Lite Algorithm}
\label{sec:algorithm}

In this section, the T$^\star$-Lite algorithm is presented for fast time-risk optimal motion planning for variable speed vehicles in obstacle-rich scenarios. Currently, there is no efficient algorithm to find an exact solution that satisfies Eq.~\ref{eq:totalcost}. An efficient method to find an approximate optimal solution to the above problem was first developed using a novel grid-based approach in T$^\star$ \cite{Song2019_TStar}. While this method provided high-quality solutions offline, the grid-based framework limited its application for on-demand real-time path planning, which is often necessary in dynamic or partially-known environments. Furthermore, when T$^\star$ was developed, the existing Wolek's kinodynamic motion model for connecting two states, required nonlinear optimization. This necessitated to form look-up tables to connect neighboring vehicle states in a grid to mitigate the high computation costs. Thus, these prior motion models could not be feasibly deployed in a rapid sample-based motion planning framework with non-uniformly spaced vehicle states.

To the best of our knowledge, our recently developed GMDM \cite{Wilson2019_GDubins} is the only kinodynamic motion model that generates high-quality multi-speed trajectories connecting any two states with a computational complexity on par with the Dubins models. This model enables the use of intermediate speeds, which lets the vehicle to select speed from a set of speeds depending on the distance to the obstacle to avoid risk, and hence enable the generation of time-risk optimal paths.

As such, the recent developments of: i) the time-risk cost function in T$^\star$ and ii) the GMDM for variable-speed vehicles enables the unique opportunity to develop the T$^\star$-Lite algorithm for fast time-risk optimal motion planning. This algorithm enables the construction of near time-risk optimal paths in a computationally efficient manner, specially under strict computation time budgets in dynamic environments. 

Figure~\ref{fig:Figure1} provides an illustrative example of the features of T$^\star$-Lite. Figure~\ref{fig:Figure1} (a) shows the key components for calculating the time-risk costs for a candidate path. Notably, the risk cost is proportional to the collision time of each of the sampled vehicle states along the path, whereas the time cost is proportional to the travel time of the autonomous vehicle along the path. Figure~\ref{fig:Figure1} (b) shows the time-risk optimal solution obtained from RRT$^*$ with the Generalized Multi-speed Dubins motion model. The autonomous vehicle operates at the maximum speed when obstacles are far away, and it slows down in the tight corridors in order to mitigate risk. The 4D states are also visualized with their headings and speeds. 

As previously mentioned, T$^\star$-Lite utilizes the RRT* framework in this paper, but without loss of generality, it can be easily implemented in other sample-based planning algorithms. The RRT* algorithm consists of six main functions: \textit{sampling, distance, nearest neighbor, near-by vertices, collision check} and \textit{local steering}. In order to bring the features of T$^\star$ using the GMDM to this framework, only the \textit{sampling, distance,} and \textit{local steering} functions are updated. All other functions remain standard. For brevity's sake, we refer the reader to \cite{Karam2010_RRT*Conference} for details on the RRT* algorithm. 

\subsection{Sampling Function}
Given a search space $A\subset\Real^2$, define $A_{free}$ as the free space and $A_{obs}$ as the obstacle region such that $A=A_{free}\cup A_{obs}$ and $A_{free}\cap A_{obs}=\emptyset$. As described before, the states $\mathbf{p}=(x,y,\theta,v)$ are sampled uniformly such that $(x,y)\in A_{free}$, $\theta\in[0,2\pi)$, and $v\in [v_{min},v_{max}]$.  For convenience, we define the set of all possible states as $\mathbf{P}$.

\subsection{Distance Function}
Let $dist:\mathbf{P}\times\mathbf{P}\rightarrow \Real^+$ be a function that returns the cost of the optimal trajectory between two feasible states, assuming the trajectory connecting the two states does not collide with obstacles. In other words, given two states $\mathbf{p}_i,\mathbf{p}_{i+1}\in\mathbf{P}$, and the optimal trajectory $\gamma_{i,i+1}^*$ connecting these two sample points, the distance is $dist(\mathbf{p}_i,\mathbf{p}_{i+1}) = J(\gamma_{i,i+1}^*)$.

\subsection{Local Steering Function}
Given two states $\mathbf{p}_i,\mathbf{p}_{i+1}$, the \textit{steer} function produces the optimal trajectory starting at $\mathbf{p}_i$ and ending at $\mathbf{p}_{i+1}$. It should be noted that $J(steer(\mathbf{p}_i,\mathbf{p}_{i+1}))=dist(\mathbf{p}_i,\mathbf{p}_{i+1})$. Finding the optimal trajectory depends on two items: i) the kinodynamic motion model used to create the candidate trajectories connecting the two states, and ii) the approximate optimization function from T$^\star$ to evaluate the time-risk costs for the generated trajectories connecting the two states.

\vspace{6pt}
\subsubsection{Generalized Multi-speed Dubins Motion Model}
The recently developed GMDM \cite{Wilson2019_GDubins} is a fundamental extension of the Dubins motion-model that enables the selection of any feasible speed for any of the three segments of the Dubins paths (i.e., $L$, $S$, or $R$). This extension gives multi-speed autonomous vehicles better maneuverability by controlling the speed, and thus turning radius, to provide faster and safer paths. It enables risk avoidance by slowing the autonomous vehicle down only near obstacles without sacrificing the travel time cost. The GMDM has a mathematical guarantee of full reachability to connect any two states in open spaces, and its closed-form solutions provide a low computational complexity. These benefits make the GMDM suitable for on-demand motion planning in obstacle-rich environments.

The equations for the GMDM are derived similarly as the Dubins model in \cite{Shkel2001_DubinsDerivation}. First, the elementary motions for straight $S$, left $L$, and right $R$ maneuvers are defined for the inertial vehicle presented in Eq.~(\ref{eq:vehiclemotion}). For the left $L$ and right $R$ primitives, we define parameter $\sigma_c$ that specifies the amount of rotation along the curve in radians between $[0,2\pi)$ to the left or right, respectively. For the straight $S$ primitive, we define parameter $\sigma_d$ to specify the distance the vehicle travels in a line. We also specify the turning rate ($u_{\sigma_c}$ for the curves) and the speed ($v_{\sigma_c}$ for the curves and $v_{\sigma_d}$ for the straight line) for the vehicle along these motions. The turning radius is therefore computed as $r_{\sigma_c} = v_{\sigma_c}/u_{\sigma_c}$ for the $L$ and $R$ primitives. 

For a particular pose and these vehicle parameters, we define the function to find the next pose for each motion primitive as $S_{\sigma_d}(x_i,y_i,\theta_i)$, $L_{\sigma_c}(x_i,y_i,\theta_i)$, and $R_{\sigma_c}(x_i,y_i,\theta_i)$. The resulting function is geometrically derived as:

\small
\begin{subequations}
\label{eq:ElementaryMotion}
\begin{align}
    \begin{split}
    \label{eq:SMotion}
    S_{\sigma_d}(x_i,y_i,\theta_i) ={}& \big(x_i+\sigma_d\cos{\theta_i},y_i+\sigma_d\sin{\theta_i},\theta_i\big)
    \end{split}\\
    \begin{split}
     \label{eq:LMotion}
    L_{\sigma_c}(x_i,y_i,\theta_i) ={}& \big(x_i-r_{\sigma_c}\sin{\theta_i}+r_{\sigma_c}\sin{(\theta_i+\sigma_c)}, \\ 
    & y_i+r_{\sigma_c}\cos{\theta_i}-r_{\sigma_c}\cos{(\theta_i+\sigma_c)},\theta_i+\sigma_c\big)
    \end{split}\\
    \begin{split}
     \label{eq:RMotion}
    R_{\sigma_c}(x_i,y_i,\theta_i) ={}& \big(x_i+r_{\sigma_c}\sin{\theta_i}-r_{\sigma_c}\sin{(\theta_i-\sigma_c)}, \\
    & y_i-r_{\sigma_c}\cos{\theta_i}+r_{\sigma_c}\cos{(\theta_i-\sigma_c)},\theta_i-\sigma_c\big)   
    \end{split}
\end{align}
\end{subequations}
\normalsize
When $r_{\sigma_c} = 1$, these elementary motions match those for the Dubins model presented in \cite{Shkel2001_DubinsDerivation}. With these motion primitives, the system of equations that characterize six GMDM path types ($LSL$, $LSR$, $RSL$, $RSR$, $LRL$, $RLR$) can be obtained, and the solutions to these equations are directly solved for. Further details of the model are provided in \cite{Wilson2019_GDubins}.

It should be noted that the speed of the first and last segment of the GMDM is determined by the high-dimensional continuous \textit{sampling} function, as these segments start and end with a sampled state. To produce a candidate trajectory, a speed for the middle segment must be selected. Here, we define discrete parameter $|v|$ which is the number of uniformly spaced speeds between $v_{min}$ and $v_{max}$ to consider for this segment. During run-time, for each path type, $|v|$ trajectories are produced, and the time-risk cost of each of these trajectories is evaluated. 

\vspace{6pt}
\subsubsection{Approximate Time-Risk Cost Function}
The approximate time-risk cost function is used to quickly evaluate the cost of a given trajectory connecting two states. The risk cost is considered constant along this trajectory. Given two states $\mathbf{p}_i,\mathbf{p}_{i+1}$ and some trajectory generated from GMDM connecting these two states $\gamma_{i,i+1}$, the risk cost and time cost can be separated. The resulting cost function from Eq.~\ref{eq:totalcost} reduces to:

\begin{equation}
    \label{eq:approxcost}
    J(\gamma_{i,i+1}) = \underbrace{\mathscr{R}(\gamma_{i,i+1})}_\text{risk cost} \times \underbrace{\int_{\gamma_{i,i+1}} \frac{1}{v(s)} \ ds}_\text{time cost $\mathscr{T}(\gamma_{i,i+1})$}
\end{equation}
Calculation of the time cost $\mathscr{T}(\gamma_{i,i+1})$ is straightforward. The risk cost is defined as the most dangerous state along $\gamma_{i,i+1}$, and is defined as the state with the least collision time to the obstacle in the direction tangent to the trajectory at that state. 

In order to determine the most dangerous state, a set of $M$ interpolated and uniformly spaced states on trajectory $\gamma_{i,i+1}$ is considered; see Figure~\ref{fig:Figure1} (a). These intermediate states are denoted as $\mathbf{\hat{p}}_{i,i+1}^m$ where $m=1,\dots,M$, $\mathbf{\hat{p}}_{i,i+1}^1=\mathbf{p}_i$, and $\mathbf{\hat{p}}_{i,i+1}^M=\mathbf{p}_{i+1}$. For each state $\mathbf{\hat{p}}_{i,i+1}^m$, the collision distance $d_{i,i+1}^m$ is determined using the autonomous vehicle's sensors. The velocity of this state is denoted as $v_{i,i+1}^m$. Thus, the collision time is computed as:

\begin{equation}
    \label{eq:collisiontime}
    t_{i,i+1}^m = \frac{d_{i,i+1}^m}{v_{i,i+1}^m}
\end{equation}

For any state, the vehicle is considered safe if the corresponding collision time is greater than a safety threshold $t^\star\in\Real^+$, which is the amount of time a vehicle needs to regain control and correct its course. Based on the collision time $t_{i,i+1}^m$ and safety threshold $t^\star$, the risk at state $\mathbf{\hat{p}}_{i,i+1}^m$ is:

\begin{equation}\label{eq:sample_pose_risk}
risk(\hat{\mathbf{p}}_{i,i+1}^m) = 
\begin{cases}
1 + \log \left({\frac{t^\star}{t_{i,i+1}^m}} \right) & \text{if } t_{i,i+1}^m < t^\star \\
1 & \text{if } t_{i,i+1}^m \geq t^\star
\end{cases}
\end{equation}
Finally, the risk cost for candidate trajectory $\gamma_{i,i+1}$ is:
\begin{equation}\label{eq:risk_function}
\mathscr{R}(\gamma_{i, i+1}) = \max_{m \in \{1,\ldots, M\}} \Big(risk(\hat{\mathbf{p}}_{i,i+1}^m)\Big)^k
\end{equation}
where $k\geq 0$ is the risk weight parameter. More information on the time-risk cost function can be found in \cite{Song2019_TStar}.

\begin{figure*}[t!]
    \centering
        \centering
        \includegraphics[width=0.8\textwidth]{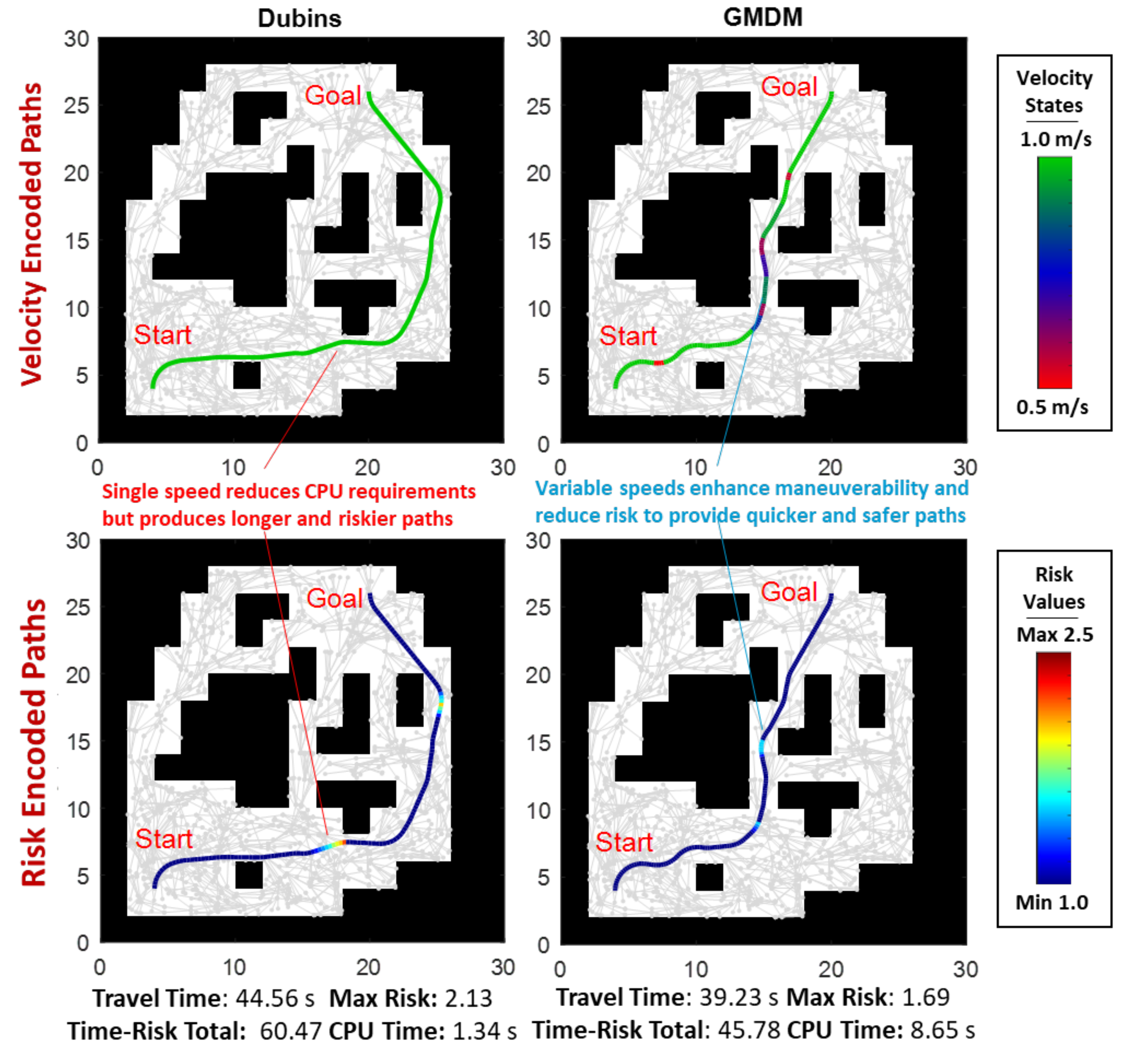}
         \caption{Comparison of the asymptotically time-risk optimal paths generated from T$^\star$-Lite for the Dubins model and the GMDM.}
         \label{fig:results}
\end{figure*}

\section{Results and Discussion}
\label{sec:results}
In this section, the asymptotically time-risk optimal paths generated by T$^\star$-Lite are presented. The scenario considered is a $30m\times 30m$ map populated with several obstacles. The autonomous vehicle has a minimum speed of $v_{min}=0.5m/s$ and a maximum speed of $v_{max}=1.0m/s$ with a maximum turning rate of $u_{max}=0.5rad/s$. This yields the turning radii $r_{min}=1.0m$ and $r_{max}=2.0m$ corresponding to $v_{min}$ and $v_{max}$, respectively, with collision time $t^\star = 6.0$ seconds and risk weight $k=2.0$. While the speeds of the start and end segments of the GMDM trajectories are determined by the continuous \textit{sampling} function, we consider up to $|v|=3$ uniformly-spaced speeds for the middle segment for each path type in order to produce the candidate trajectories. In order to calculate the risk cost, $M=4$ interpolated states are generated on each candidate trajectory. Then, the time-risk cost function is used to select the optimal GMDM trajectory connecting two states. The search tree was configured to have up to 3000 randomly sampled states. Up to 100 nearest-neighbors are considered when updating the search tree with a newly sampled state. The maximum connection distance between sampled states was set to 3 meters.  In order to show the utility of variable-speeds in time-risk motion planning using the GMDM, results were compared with a maximum-speed Dubins vehicle. The experiment was carried out in MATLAB using the RRT* path planner in the Navigation Toolbox on a Windows 10 machine with an Intel Core-i7 7700 CPU and 32GB of RAM.

Figure~\ref{fig:results} shows the results generated by T$^\star$-Lite by using both the Dubins motion-model (left column) and the GMDM (right column). Each plot shows the time-risk optimal paths for the corresponding motion model. The first row shows the velocity states on the time-risk optimal paths, and the second row shows the associated risks encoded on the paths. 

With the maximum-speed Dubins vehicle, T$^\star$-Lite determines that the best path avoids the obstacle-dense regions in the center in order to minimize the number of risky high-speed maneuvers, and thus selects a long and indirect route. Due to its fixed high-speed state, high risk is unavoidable as the vehicle navigates through the corridor. It is noted that utilizing the sample-based framework allows the vehicle to tightly wrap around the obstacles in a continuous manner instead of taking several right-angle turns in the grid-based approach of T$^\star$.  

The GMDM vehicle, on the other hand, is able to utilize variable speeds effectively and efficiently in order to safely navigate through the obstacle-rich region in the center of the map while noticeably improving the time-risk cost. This is achieved since i) the multiple speeds, and thus multiple turning radii, allow for tighter maneuvers around obstacles and ii) the optimal speeds are selected based on obstacle distance in order the minimize the joint time-risk cost. As a result, the travel time cost is improved by over 5 seconds and the maximum risk is substantially reduced when compared to the Dubins vehicle. Overall, the GMDM provides superior time-risk optimal paths for variable-speed autonomous vehicles at only a slightly increased computational cost.

\section{Conclusions and Future Work}
\label{sec:conclusions}

In this paper, we develop a new algorithm, called T$^\star$-Lite, for rapid time-risk optimal motion planning for multi-speed autonomous vehicles in hazardous and dynamic environments. This is achieved by i) porting the novel time-risk cost function from our previously developed T$^\star$ algorithm into the asymptotically-optimal RRT* framework and ii) utilizing the Generalized Multi-speed Dubins Motion-model to provide near-optimal trajectories connecting waypoints in a computationally efficient manner. Specifically, T$^\star$-Lite is able to select the asymptotically optimal speeds, depending on the vehicle distance to obstacles, that jointly minimizes the time-risk cost. This results in trajectories that enables the autonomous vehicle to quickly but safely maneuver through obstacles-rich regions and thus follow a more direct path to the goal, yielding shorter travel times with substantially reduced collision risk. 
The simulation results validate our claim by showing that high-quality paths considering both time and risk for multi-speed vehicles are quickly obtained. 

Future work includes an in-depth analysis of the T$^\star$-Lite framework using other asymptotically-optimal sampling-based frameworks like PRM* and its derivatives. Direct comparisons will also be made with the grid-based T$^\star$ algorithm in terms of both computation time and solution quality. Smart high-dimensional sampling methods will also be developed to limit the size of the search space to improve both computational performance and solution quality. These methods will be informed based on the autonomous vehicle characteristics, decision variables, and the topography of the environment, using both model-based and data-driven approaches.  Finally, new kinodynamic motion models will be developed that can quickly produce high-quality trajectories connecting waypoints while considering the full state information of the autonomous vehicle, such as acceleration and jerk, to ensure smoother autonomous vehicle operation. The time-risk optimal cost planning could be extended to multi-agent resilient systems~\cite{SG19}.

\bibliographystyle{ieeetr}
\bibliography{References}

\end{document}